\RequirePackage{fix-cm}
\documentclass{scrartcl}
\usepackage{color,soul}
\usepackage{marvosym}
\usepackage[affil-it]{authblk}
\usepackage{amsmath}
\usepackage{setspace}
\usepackage{amssymb}
\usepackage{appendix}
\usepackage{graphicx}
\usepackage[backend=biber, natbib, style=authoryear, citestyle=authoryear]{biblatex}
\begin{document}

\title{Making AI Meaningful Again }


\author{Jobst Landgrebe
  \thanks{Electronic address: \texttt{jobst.landgrebe@cognotekt.com}; Corresponding author} }
\affil{Cognotekt GmbH, Bonner Str. 209, D-50996 K\"oln, Germany}

\author{Barry Smith
  \thanks{Electronic address: \texttt{phismith@buffalo.edu}}}
			\affil{University at Buffalo, Buffalo, NY, USA} 

\date{\today}

\maketitle

\begin{abstract}
Artificial intelligence (AI) research enjoyed an initial period of
enthusiasm in the 1970s and 80s, but this enthusiasm was tempered by a long
interlude of frustration when genuinely useful AI applications failed to be
forthcoming. Today, we are experiencing once again a period of enthusiasm,
fired above all by the successes of the technology of deep neural networks
or deep machine learning. In this paper we draw attention to what we take
to be serious problems underlying current views of artificial intelligence
encouraged by these successes, especially in the domain of language
processing. We then show an alternative approach to language-centric AI, in
which we identify a role for philosophy.

\end{abstract}

\section{The current paradigm of AI: Agnostic Deep Neural Networks (dNNs)}
\label{sec:1}

An AI application is a computer program that can create an output in
response to input data in a way that is similar to the
ways humans react to corresponding environmental stimuli. In what follows
we will focus on AI applications that work with natural language input, where the
currently dominant paradigm is provided by what is called agnostic deep
machine learning.\footnote{Also referred to as `brute force' learning} The
latter is a subfield of applied mathematics in which
input-output-tuples of data are used to create stochastic models, in a
process often (somewhat simplistically) referred to as `training'. The
inputs are connected to outputs probabilistically, which means that there
is a certain (a priori unknown but measurable) likelihood that a given
input will be associated with a given output. The models are referred to as
`stochastic' because they work by utilizing the fact that the data on which
they draw is probabilistic in this sense.
The models are, in addition, `agnostic'  ~--~  which means that they do not
rely on any prior knowledge about the task or about the types of situations
in which the task is performed, and they are often ``end to end,'' which
means that they are meant to model an entire process such as answering a
letter or driving a car. The models are, finally, `deep' in the sense that
their architecture involves multiple layers of networks of computational
units (thus not, for example, because of any depth in their semantics) 

For agnostic deep learning to be useable in creating an AI application, a
number of conditions must be satisfied:

\begin{enumerate}
\item	A sufficient body of training data must be available in the form of
tuples of input and output data. These are digital mappings of,
respectively, a situation in response to which an action is required, and
an action of the corresponding sort \citep{hastie:2008}. A classical
AI-application in this sense is the spam filter, whose initial output data
were created using annotations, in this case adding the label ``spam'' to
email inputs.
\item	Computers must receive the training material 
in digital form, so that it can be processed using the computing
resources available today \citep{cooper:2004}.
\item	The annotated training tuples must be reasonably consistent
(noise-poor)  ~--~  that is,
similar inputs should lead to similar outputs. This is because machine
learning requires repetitive patterns  ~--~  patterns that have arisen in a
recurring, rather than erratic, process. The behaviour of human email users
when identifying spam forms a repetitive process of the needed sort. The
reason for this is that users of email have a motive to become experts in successful
identification of spam, since they are aware of the high costs of failure.
The movement of the oil price over time, in contrast, is an example of an
erratic process.  This is because the input data pertaining to
geopolitical and economic events bear no consistent relation to the
output data, for example the price of Brent crude.
\item	The data input must be abundant, since a machine-learning algorithm
is a stochastic model that needs to represent as much as possible of the variance which
characterises the situation in which the model is to be used. Because in
language applications the overall complexity of the relationship between
input and output is typically very high, the models will need many
parameters. For mathematical reasons these parameters can only be
estimated
(through the type of optimisation process otherwise called ``training'') on
the basis of huge data sets. If the training sets are too small, there is a
high chance that novel input data will not have the properties of the data
sampled in the training distribution. The model will then not be able to
produce an adequate output under real production conditions.
\end{enumerate}

Most of the AI applications in current use, for example in product
recommendation or advertisement placement, draw on situations where these
four conditions are satisfied.
To establish the training set for the first spam
filters, developers needed to collect millions of input-output data tuples
where inputs are emails received by humans and outputs are the
classifications of these emails by their respective recipients either as
spam or as valid email. They then trained a machine-learning model using
these data tuples and applied the result to new emails. The goal is that the
model should replicate the human reaction it has been trained with, which
means: identifing spam in a way that matches, or even outperforms, the
behaviour of a typical human.

In classification applications such as this, it is only knowledge of a very simple type
~--~ knowledge captured by simple input-output-tuples
~--~  that is given to the machine by its mathematician or AI-engineer
trainers. In some cases, application developers may wish to improve the model
that is generated by the algorithm from the data by selecting for training
purposes only those tuples that have certain desired properties (as when,
in building training
models for autonomous cars, they in effect seek to approximate the
positive aspects of the driving behaviour of mature females
rather than that of teenage males). The performance of the
machine is in such cases designed to surpass that of the average human, because
the trainers of the model select only the most desired sorts of responses
from what may be a much more considerable variance exhibited in actual
behaviour. Designers may also select data that have been somehow
validated by experts for correctness, creating what is called a ``gold
standard'' set of annotations. Because the engineer uses prior knowledge
about data quality when making such selections, this is equivalent to an  ~--~ 
albeit minimalistic  ~--~  usage of prior knowledge in machine learning.

When such strategies are followed, machine learning with neural networks
can out-perform even the strongest
human performance with regard to both efficiency and
effectiveness,\footnote{Increasing efficency means: reducing unit
production costs; increasing effectiveness means: achieving higher desired quality
per production unit.} and we can now distinguish three types of cases in which
such better-than-human performance is achievable:

\begin{enumerate}
 \item \textit{dNNs with higher efficiency than is obtainable by
 humans}:
 when the behaviour that is modelled consists of truly repetitive
 processes with narrow scope and with data that can easily be represented
 in digital form: for example in complex industrial
 automation tasks, following a pattern that has been the driver of
 engineering since the industrial revolution.
 \item	\textit{dNNs with higher effectiveness than humans}: in
 hypothesis-based pattern identification, for example in the
recent identification by a dNN of a correlation between retinal patterns
and cardiovascular risk factors \citep{poplin:2018}. 
\item \textit{AI with both a higher efficiency and effectiveness than
with humans}: achieved in reinforcement learning, a method used in certain
narrowly defined situations of the sort that arise in games (for example in
GO \citep{silver:2016} or First Person Shooters \citep{jaderberg:2018}) and
in contexts that can be framed like games \citep{sutton:2018}. 
\end{enumerate}

Examples of applications under each of these headings that will be
possible in the near future are: (i) driving a car on a
highway under (near-)average weather conditions, (ii) scientific pattern-search
applications, for example in biology or astronomy, (iii) maintenance
robotics as in industrial nuclear plant waste removal.

Unfortunately, each of these types of situations is highly restrictive,
and none occurs where we are dealing with natural language input.

\section{Applying agnostic Deep Neural Networks in the field of language understanding}
\label{sec:2}

To understand how modern agnostic deep-neural-network AI works in the
language domain, consider the most prominent production example, which is
that of machine translation as illustrated by Google
Translate.\footnote{https://translate.google.com/} A recent publication
authored by Google
Brain\footnote{This is the official name of Google's AI department. While
Google's machine-learning engineers are certainly among the world's leading
representatives of their craft, the name nonetheless reveals a certain
hubris.} and Google Research with the title ``Attention is all you need''
\citep{vaswani:2017} provides a representative example. The stochastic
models described in this paper were trained for the translation of English
to German and of English to French. To train \textit{Transformer} ~--~  which is the
best-performing ``big'' model described in the paper ~--~  the authors encoded
the language material at their disposal using the method of byte-pair encoding, which
encodes each single-sentence input into an encoding vector of 1024 real
numbers (rounded to a certain number of decimal places\footnote{This
encoding approach is used (with variations on how the vector is created) by
all dNNs since ``word2vec'' \citep{mikolov:2013}.}). This is a
complexity-reducing encoding, which means (very roughly) that it treats
each sentence simply as a series of meaningless
signs.\footnote{\label{fn}In this entire text, ``meaning'' signifies
the relevance to the actions and thoughts that humans attribute to the
stimuli that they encounter in sensation. For a non-English speaker, an
English sentence, too, is a series of meaningless stimuli. For
an English speaker, in contrast, the sentence is immediately interpreted as
meaningful.} This allows the encoding process
to retain certain important features of the input sentences because
relevant sentence patterns are repeated in many sentences in a similar way,
and these sentences are shown to the algorithm.\footnote{For example, the
algorithm learns to translate the German word `Mehl' into `flour' because this
pair is repeated many times in training sentences. But it will fail to
translate ``Wir haben Mehl Befehl gegeben zu laufen'' into the adequate ``We
ordered Mehl to run''. It rather gives out the nonsensical ``We have ordered
flour to run'' (result produced on Jan. 7, 2019). The translation fails
because there are not enough training examples to learn the martial usage
of surnames without title.}

But at the same time, it necessarily leads to the discarding of many
subtleties of these sentences. This is because the embedding of the
sentence loses relations not only between words within the sentence but
also between sentences. For a further feature of the experiments reported is
that the models used are trained with quite small amounts of training data:
36 million sentence pairs for English-French, and only 4.5 million for
English-German. The models are completely agnostic: they have no knowledge
of linguistics, for example, because they have no knowledge of anything at
all. Rather, they just try to mimic the human translations (or rather the
corresponding sets of vectorially simplified input-output pairs) they
learned from. The principal problem with this approach, however, is that
embedding into a linear vector of encoding real numbers ~--~ no matter
how long the vector is ~--~ leads to the discarding of all
information pertaining to the contexts of the input sentences. That this
has adverse consequences becomes clear when we reflect that, in all
language interpretation processes, even for single sentence inputs, humans
use prior knowledge to contextualise the sentences they receive. As an
example, consider how a typical reader of this text would contextualise the
single sentence: ``In the beginning was the word.''\footnote{To a
reader without knowledge of the Bible this sentence (John, 1,1) will seem
strange or unintellegible. It is impossible to enumerate all
such contextual constellations and include them as annotated features to
training sets for stochastic models in amounts sufficient for
machine learning.} 

\subsection{Results thus far}
\label{subsec:2.1}
How well, then, do these models do? \textit{Transformer}, specifically, creates a
model that achieves a sentence-level score of 28.4 for English-German and
41.8 for English-French using the BLEU metric, which measures on a scale
from 0 to 100 the degree of matching of the machine-translation with a
human gold-standard translation \citep{papineni:2002}. A
score of 100 can never be achieved because there are always several valid
translations for any given sentence and not all of them can be in the
gold-standard set. But 75-85 could be achieved in theory. Such a score
would be excellent, and it would correspond to the translation abilities of
an average bilingual speaker. The scores achieved by \textit{Transformer}, in
contrast, which are reported as the state-of-the art in machine
translation, are low.\footnote{To illustrate the limitations of the approach,
Hofstadter used input sentences with a high degree of
cross-contextualisation (see ``The Shallowness of Google Translate'',
\textit{The Atlantic}, January 30, 2018).

Text by Hofstadter: \textit{In their house, everything comes in pairs. There's his
car and her car, his towels and her towels, and his library and hers.}

Google Translate: \textit{Dans leur maison, tout vient en paires. Il y a sa voiture
et sa voiture, ses serviettes et ses serviettes, sa bibliothèque et les
siennes.} 

Translated back into English by Google: \textit{In their house everything comes in
pairs. There is his car and his car, their napkins and their napkins, his
library and their's.}} The reason for this shallowness of this
so-called ``neural machine translation'' is that the vector space it uses
is merely morpho-syntactical and lacks semantic dimensions.

\subsection{General limitations of machine learning}
\label{subsec:2.2}
Major limitations of current deep-learning paradigms have been identified
already (for example in \citep{marcus:2018}). They include first of all a
set of quite general problems affecting stochastic models of any sort ~--~  not
only deep neural nets but also traditional regression and classification
approaches \citep{hastie:2008}, including graph-based stochastic models (Bayesian Networks).

The first of these limitations turns on the \textit{huge data need} of stochastic
models, which may employ millions of parameters. \textit{Transformer}, for example,
has 213 million para\-meters and needs at a minimum billions of data tuples
to become useful even for the sorts of rough translation produced by Google
Translate. This limitation is already of considerable importance given
that, leaving aside the resources of internet giants such as Google, there
are few real-world examples of data available in the sorts of quantities
needed to deal with complex outcomes using any sort of stochastic approach.

Second, all stochastic models require a \textit{stable environment}. The quality of
their output depends on how well they reflect the real-world input-output
relationship they are aiming to represent. Where this relationship is
erratic, there can be no good model (consider again the oil price example
mentioned above).
But even where the relationship is stable, the model will quickly become
invalid if the input-output relationship changes on either side even in
some minor way. This is because the model does not generalise. Once fed
with data as input that do not correspond to the distribution it was
trained with, the model will fail. And it will not alert the user, because
it will not know that it is
failing.\footnote{So-called deterministic AI models \citep{russell:2014} do
not generalize, either, but they report their failures.} This explains why
stochastic spam filters and
similar applications are so vulnerable to changing situations, and why they
so often need re-training. And the more complex an application, the more
demanding will be the re-training of its network that is required upon
change of input constellations (for example when new types of
sensors are introduced in driverless cars). The costs for such re-training
will vary, of course, with the complexity of the input and the accuracy
requirements of the network.

But there is a third group of limitations, turning on the fact that
\textit{the output of all stochastic models is, by definition,
approximative.} Models of
this sort can yield only the most probable output for any given input and
model, and this output often falls below even the average human output. For
many imaginable useful purposes, however, the output should be at
least as reliable as the behaviour not of the average but of a qualified
subset of human reference samples. This is very hard to achieve in
language-focused applications using dNNs only.
Unlike machines, humans are able
spontaneously and immediately to attribute meaning to the world they
experience. This is because the human species has evolved with a complex
set of dispositions to react immediately in highly specific ways to
specific sorts of external stimuli.

Human beings are, along many
dimensions, tuned to the environments in which they live. The entities that
we experience are spontaneously assigned meanings that
reflect their relevance to our survival, meanings that are assigned using
mechanisms hardwired into our brains. The belief that
stochastic models can learn to make decisions without benefit of prior
hardwiring of this sort is as naive as the old tabula rasa theories that
were once the staple of empiricist philosophers and of their empirical
psychologist followers. Such views were criticised by \citet{gibson:1979} in his
ecological theory of perception\footnote{Indeed they
were criticised, 200 years earlier, by Immanuel Kant in 1781 in his
\textit{Critique of Pure Reason}.}, and they were experimentally refuted in
the works on infant cognition of \citet{carey:2001}, \citet{gopnik:2000},
\citet{keil:1989}, \citet{keil:1995}, \citet{kimSpelke:1999}, who
demonstrated that infants ~--~ and primates \citep{povinelli:2000}) ~--~ possess a large
body of categorical and structural knowledge about the world of solid
objects long before they even start acquiring the grammar of their mother
tongue \citep{leslie:1979}.
Indeed, it seems that language acquisition presupposes the working of a
common set of ontological distinctions on the side of language learners,
including the distinction between objects and processes, between
individuals and categories, between natural and accidental properties of
objects, and so forth.

Even the theory of Bayesian models for concept learning based on
similarity acknowledges (i) the need for a prior genus-individual
distinction to explain the mechanics behind generalization and (ii) the
existence of a prior meta-heuristic linking membership in a class to
property instantiation \citep{tenenbaum:1999, tenenbaum:2001}. As
\citet{rehder:1999} formulates the matter, categorization relies on inferences about the
causal role of putative essences in producing observable features. Such
features are, in other words, merely secondary,
derivative; and all the naive knowledge brought to bear by the infant
follows from the natural and universal supposition that things belong to
classes sharing similar properties \citep{medin:1989, solomon:1999}. Even
children as young as 3 believe that the `insides' of objects are
relevant in determining class membership \citep{gelman:2003, gelman:1991a,
keil:1989}. According to \citet{carey:2001} (p. 207),
experiments on object recognition suggest that there is an
object tracking system in the infant ~--~  a system that tracks
three-dimensional, bounded, and coherent physical entities, and fails to
track perceptually specified figures that have a history of non-cohesion.
And what holds of infant cognition in general holds also of infant language
learning and language competence in particular, where the capability of
object tracking grounds the use of nouns and pronouns. Indeed, part of the
background source of this empirical work on infant ontology was Chomsky's
work \citep{chomsky:1956} on innate universal grammar. 
\citet{gelman:1991} make explicit reference to these ideas when
they assert that they are able to ``determine how languages and conceptual
systems are constrained by examining the forms and meanings that children
construct, and which errors they fail to make'' (\citet{gelman:1991};
compare \citet{millikan:2001}, p. 47).

For our purposes here, it is crucial that the AI applications running on
today's computers can simulate at best only small fragments of the
hard-wired human capabilities revealed in such research. This means that
they can simulate only small fragments of the semantics underlying human
language use. As we shall see, neural networks do in this respect 
have limitations no less severe than those of traditional logic-based AI approaches to the
modeling of human cognition. The formal ontologies used in
the latter, however, allow us to overcome some of the limitations of the
former because they involve 
direct representations of the sorts of objects,
processes and attributes (and associated nouns, verbs and predicates) used
by human beings in perceiving, acting and speaking. The training of neural
networks is the attempt
to build simulacra of relation-rich content of this sort out of gigantically large
numbers of features represented using numerical input vectors or matrices.
The training algorithms estimate what amounts to a very large polynomial
(this is what a neural network is) with the help of an optimization
procedure. But building the simulacra in this way seems to be infeasible even for simple ontologies of
the RDF-sort made up of structures of the type: \textit{entity A -- relates to --
entity B} \citep{gutierrez:2018}. 

\subsection{Limitations applying specifically to deep neural networks (dNNs)}
\label{subsec:2.3}

As humans process sensory input data, they assign meanings to the objects
and events which the data represent (and from which the sensory content
originates), experiencing these objects and events as belonging to a
specific sort of categorical structure. dNNs, in contrast, do not use any of the
target-derived properties of the input data that humans spontaneously use
when they assign meaning to the data which they receive through experience.
The result is a tremendous brittleness of dNN capabilities.
\citet{moosavi:2017} describe how high-performance neural networks
developed for image classification can be nudged into a complete
misclassification of images when the input material is mixed with a perturbation
image. For example, what is at first correctly classified by the system as
a \textit{flagpole} is classified as a \textit{labrador} after the system
is very slightly
perturbed. Perturbations of an analogous sort do not cause problems for
humans at all. As \citet{jo:2017} showed, dNNs used in image recognition work merely by
learning certain surface-statistical regularities from images: the green
grass that forms the typical background of a cow, for example, is
contrasted with the grey of asphalt that forms the typical background of a
car. They can be perturbed so easily precisely because they do not learn
what the images are about and the sort of world to which the imaged objects
and events belong. 

The same holds also of the dNNs constructed for language processing
purposes. A recent paper by \citet{chen:2017}
\textit{proves mathematically} (which, given what was said above, we should in any case expect)
that dNNs lack core computational features of traditional approaches
to syntactic language analysis, for example, of the sort pioneered by
\citet{chomsky:1956} using probabilistic
context-free grammars. As the authors show, while it is
required of every valid stochastic model that it compute a valid
probabilistic distribution, this condition is not in general satisfied by
dNNs working from language input. But without this ability, there can be no
computational representation of semantics. Thus, as shown in
\citet{feng:2018}, the language constituents used by dNNs to make
predictions in
question-answering or textual entailment tasks often make no sense to humans at
all.\footnote{One example described in \citet{feng:2018} 
rests on the input: ``In 1899, John Jacob Astor IV invested \$100,000 for
Tesla to further develop and produce a new lighting system. Instead, Tesla
used the money to fund his Colorado Springs experiments''. The described
system correctly answers the question: ``What did Tesla spend Astor's money
on?'' with a confidence of 0.78 (where 1 is the maximum). The problem is
that it  provides exactly the same answer with a similar degree of
confidence as its response to the nonsensical question: ``did?''}

This in turn means that dNNs, whatever it is that they are doing, cannot
be modeling the semantics that need to be captured in order to extract
information from texts, a crucial task in natural language processing for
most automation purposes.

The information extraction (IE) results presented in \citet{zheng:2017}
provide a poignant example of the low quality currently being achieved for
tasks of this sort.\footnote{The $F_1$-score of 0.52 reported by \citet{zheng:2017} seems
quite high; but most of the training material is synthetic and the reported
outcome only concerns information triples, which cannot be used for applied IE.
The example is `poignant' because the paper in question won the 2017 Prize for
Information Extraction of the Association for Computational Linguistics,
globally the most important meeting in the language AI field.} This
example reveals just how low the expectations in the field have become. The failure
of dNN-based approaches to compute natural language semantics is
illustrated also by the recent misclassification of the United States
Declaration of Independence as hate speech by the Facebook filter
algorithms.\footnote{https://www.theguardian.com/world/2018/jul/05/facebook-declaration-of-independence-hate-speech}

dNNs are also unable to perform the sorts of inferences that are required
for contextual sentence interpretation. The problem is exemplified by the
following simple example:

\begin{center} ``The cat caught the mouse because it was slow'' vs.\\``The cat
caught the mouse because it was quick.'' \end{center}

What is the ``it'' in each of these sentences? To resolve anaphora requires
inference using world knowledge ~--~  here: about persistence of object identity,
catching, speed, roles of predator and prey, and so forth. Thus far,
however, little effort has been invested into discovering how one might
engineer such prior knowledge into dNNs (if indeed this is possible at
all).\footnote{Currently, prior knowledge is used mainly for the selection
or creation of the training data for end-to-end dNN applications.} The result is that, with the
exception of game-like situations in which training material can be
generated synthetically, esp.  in reinforcement learning, dNN models built
for all current
applications are still very weak, as they can only learn from
the extremely narrow correlations available in just that set of annotated
training material on the basis of which they were created. Even putting
many dNN models together in what are called ``ensembles'' does not overcome
the problem.\footnote{The improvements provided by this approach are
very modest and not higher than those achieved by other tweaks of dNNs such as optimised
embeddings or changes in the layering architecture} \citep{kowsari:2017}.

And worse: because the dNNs rely exclusively on just those correlations,
they are also unable to distinguish \textit{correlation} from
\textit{causation}, as they can
model only input-output-relationships in ways that are agnostic to
questions of, for example, evidence and causality. Thus they can detect that
there is some sort of relationship between smoking and lung cancer. But
they cannot determine the type of relation that is involved unless
references to this very relation and to relevant types of relata themselves form
part of the annotated corpus. Unfortunately, to create the needed annotated
gold-standard corpora ~--~  one for each domain of interest ~--~  would be hugely
expensive in terms of both time and human expertise. To make dNNs work
effectively in language applications thus would require not only enormous collections of
data but also, at least for many applications ~--~  and certainly for those
examples involving the tracing of causality ~--~  the investment of
considerable amounts of human expertise.

One final problem resulting from the properties of dNNs as very long
polynomials is a lack of transparency and ~--~ in contrast to determinstic
algorithms ~--~ a black box operation mode. Therefore,
dNN engineers cannot tell why the network yielded its output
from a given input. This poses a major challenge in areas
where we need to reproduce or analyse
the behaviour of the network, for example in case of disputes over
liability.

Taken together, these problems rule out entirely the use of machine
learning algorithms \textit{alone} to drive mission-critical AI systems
~--~  for example with capability such as
driving cars or managing nuclear power stations or intensive care
units in hospitals. They are too brittle
and unstable against variations in the input, can easily be fooled, lack
quality and precision, and fail completely for many types of language
understanding or where issues of liability can arise. Even at their very
best, they remain approximative, and so any success they achieve is still,
in the end, based on luck.

\section{Making AI meaningful again}
\label{sec:3}

\subsection{Adding semantics to automation solutions}

To overcome these problems, ways need to be found to incorporate prior
knowledge into the AI algorithms. One attempt to do this is to enhance
Bayesian Networks with an explicit relationship semantics
\citep{koller:2009}, which allows the model designer to build in knowledge
describing entity relationships before using data to train the weights of
these relationships. This reduces the learning effort on the part of the
system by providing a
rudimentary form of prior knowledge. But unfortunately, the expressivity of
the resulting models is too low to represent the sorts of complex contexts
relevant to human language understanding. Furthermore, they are not exact,
secure, or robust against minor perturbations. They are also not
transparent, and thus humans
cannot reliably understand how they work to achieve given results. The goal
of meeting this requirement is now dubbed ``explainable AI'', and we
will describe one promising strategy for achieving this goal that involves
building applications that work in accordance with the ways humans
themselves assign meaning to the reality that surrounds them. 
To achieve this end, we use a semantics-based representation that is
able to deal with language as it is actually used by human beings.
Importantly, the
representation is able to incorporate prior knowledge based on low
to medium amounts of input material of the sorts found in typical
real-world situations. For humans in such situations find meaning not in
\textit{data}, but rather \text{in the objects and events that surround
them,} and in the
affordances that these objects and events support \citep{gibson:1979}. This
implies a different sort of AI application, in the building of which not
only mathematics and computer science play a role, but also philosophy.

Part of what is needed is to be found already in the early attempts to create 
`strong' logic-based AI.\footnote{An excellent
summary can be found in \citet{russell:2014}.} For our purposes here, the
most interesting example of an attempt of this sort is in the work of Patrick Hayes, a
philosopher who first made his name with a paper co-authored with John
McCarthy, commonly accredited with having founded the discipline of AI
research. The paper is titled ``Some Philosophical Problems from the
Standpoint of Artificial Intelligence'' and it lays 
forth for the first time the idea behind the calculus of situations
\citep{mcCarthy:1969}. In subsequent years Hayes set forth the idea
of what he called `na\"ive physics', by which he meant a theory consisting
of various modules called `ontologies', that would capture the common-sense
knowledge (sets of common-sense beliefs) which give humans the
capacity to act in and navigate through the physical world \citep{hayes:1985}.
The theory is axiomatised using first-order logic (FOL) and Hayes proposed
that something of the order of 10,000 predicates would need to be encoded
in FOL axioms
if the resulting theory was to have the power to simulate human reasoning
about physical objects of the sorts that are encountered by humans in their
everyday lives.\footnote{Hayes' conception of an ontology as the
formalization of our knowledge of reality continues today in the work of
Tom Gruber, whose Siri application, implemented by Apple in the iPhone, is
built around a set of continuously evolving ontologies representing simple
domains of reality such as restaurants, movies, and so forth.}
The problem with Hayes' approach, as with strong AI in general,
is that to mimic even simple human reasoning in real time would require a
reasoning engine that is decidable, and this implies a severe restriction
on the expressiveness of the logic that can be used. Standardly, one ends
up with a very weak fragment of FOL such as that encapsulated nowadays in
the so-called Web Ontology Language (OWL). OWL is restricted, for
example, in that it can capture at most relational information involving
two-place relations, and it has a similarly diminished quantifier syntax
and a well-known difficulty in dealing with time-series data.
For this and many other reasons, logic-based systems have rarely reached the
point where they were able to drive AI-applications. They did, however, spawn
the development of a huge body of mechanical theorem-proving tools
\citep{robinson:2001}, and they contributed to the development of modern
computational ontologies, which helped to transform biology into an
information-driven discipline \citep{ashburner:2000}. Both of these
developments are essential for the sort of AI applications combining formal logic and stochastic
models that we describe below.


\small
\begin{table}
\caption{Minimal desiderata for a real-world AI language processing system}
\label{tab:1}       
\begin{tabular}{p{25mm}p{65mm}p{55mm}}
\hline\noalign{\smallskip}
Property&System&Example\\
\noalign{\smallskip}\hline\noalign{\smallskip}
Exactness&needs to be able to be exact where necessary and not always
restricted to the merely approximative&in the insurance domain: automated
validation and payment of a claim\\[3mm]
Information security&needs to avoid insecurities of the sort which arise, for example,
when even slight perturbations lead to drastically erroneous outputs&in
autonomous driving: avoid harmful consequences of adversarially manipulated
traffic signs\\[3mm]
Robustness&needs to be able to work reliably in a consistent way even given
radical changes of situation and input, or to detect critical changes and
report on its own inability to cope&in any domain: content not understood by the
system is marked for inspection by a human; an alert  can be generated if
necessary\\[3mm]
Data\newline parsimony&needs to be trainable with thousands to millions of data
points (rather than billions to trillions ~--~  magnitudes which rarely occur
in reality)&in the domain of business correspondence:  automation of
letter-answering on the basis of just a few thousand examples per class of
letter\\[3mm]
Semantic fidelity&needs to be able to incorporate contextual
interpretations of input situations&in sentiment analytics:
the Declaration of Independence should not be classified as hate
speech\\[3mm]
Inference&needs to be able to compute the consequences of given inputs in a
way that allows the system to 
distinguish correlation from causality (thus requiring the ability to
reason with time and causation) &in medical
discharge summaries: determination of required actions on the
basis of text input, for example in automated processing\\[3mm]
Prior \newline knowledge \newline usage&needs to be able to use prior knowledge to interpret
situations&in claims management: understanding that issuing a declaration of inability to pay
implies earlier receipt of a payment request\\[3mm]
\noalign{\smallskip}\hline
\end{tabular}
\end{table}
\normalsize

\subsection{Inserting philosophy into AI}
\label{subsec:3.1}

\subsubsection{Desiderata for automated language processing}
We will show in what follows how, by augmenting stochastic models
(including dNNs) with philosophically driven formal logic, we can create AI
applications with the ability to solve real-world problems. We present an
example of such an application and describe how the machinery proposed is
already in commercial production. First, however, we give details of what
we take to be the minimal requirements which any real-world AI system must
satisfy (Table 1). These requirements cannot be satisfied by agnostic
machine-learning systems alone, as they presuppose the ability to deal with
the semantics of human (natural) language. They can be satisfied, we
believe, only by combining stochastic inference components with methods
associated with traditional, logic-based AI in such a way as to allow
incorporation of prior knowledge.

On the approach we shall describe, all the desiderata listed in Table~1 
are satisfied on the
basis of a formal representation of prior knowledge
using a computable representation of the natural language semantics of the
information the system is processing. To succeed, this 
representation needs two major elements: (a) a set of logical formalisms,
constituted by formal ontologies that enable the 
storage and manipulation of language in Turing machines, and (b) a
framework which enables one to define the meanings of the elements of the
language. 

We can describe only the rough outline of these components here, though
one important feature, the methodology for development and use of
ontologies to which we appeal, is described in detail in \citet{arp:2015}. 

\subsubsection{Representing natural language}
Natural language as input is of course very hard to
express in a logical framework, and a typical basic pipeline, which 
sacrifices a considerable part of the semantics in order to achieve
computability, comprises the following elements:
\begin{enumerate}
\item morphological and syntactical error correction of an input text using
dNN-models trained using large public data resources,
\item  syntactical parsing with the help of a
stochastic parser, e.g. a conditional random field parser as described in
\citet{finkel:2008}, and
\item inference applied to the parser outputs with the help of (computable)
propositional logic.
\end{enumerate}

However, to capture the semantics of full natural language, a
much stronger, higher
order intensional logic is required for its computational representation
\citep{gamut:1991}, and a logic
of this sort cannot be used for computational purposes. To enable
computation, the source text must thus be expressed by
combining several computable logical dialects which together provide a representation
that is adequate for a given context and purpose. For example,
fundamental language constructs can be represented using FOL, while temporal
relationships require temporal propositional
logic. Intensional sentences can be represented using modal logic.
A suite of such logical formalisms is able to achieve a
good approximation to natural language semantics while still allowing
computability.
Computability does not, be it noted, require decidability,
but robustness and completeness (in the technical logical sense) are
essential. Decidability is not required because logical inference is fully
possible with robustness and completeness alone. The absence of
decidability certainly
implies that on occasion a computation may not terminate. To take account of
such cases, however, algorithms are stopped after a
pre-determined maximum computation period whose length is defined by the
qualitative requirements of the system. Such cases can then be
handed to a human for decision.

The resulting system does not, however, strive for general AI. It
always works in specific sub-domains covering segments of reality. The execution of the
logics is then orchestrated with the
help of domain-specific deterministic or stochastic controllers which are
needed to ensure that inference steps are carried out in the order that is
appropriate for the problem at hand.

\subsubsection{Digression: Ambiguity and indeterminancy} \label{unc}
A hybrid AI system along the lines described can also deal with phenomena such as ambiguity,
vagueness and indeterminancy, which natural language frequently contains.
We cannot give a full account of how these phenomea are dealt with here,
but we provide two informative examples:

\begin{enumerate}
 \item Ambiguous sentences, for example ``Every man loves one woman'',
 which can mean that every man loves a woman, e.g. his wife (on the
 \textit{de dicto}
 interpretation), or that every man loves one well-known woman, for
 example Marilyn Monroe (on the \textit{de re} interpretation). To account for
 the ambiguity, an adequate logical representation creates two logical
 phrases from this sentence, and deduces the correct meaning 
 from the context, for example using stochastic models.
 \item Uncertainty, such as ``John's father did not return.
Now John is searching for him.'' Here, the transitive verb in the second
sentence may or may not have an existing object. This
phenomenon can be addressed using intensionality
tagging for certain transitive verbs and subsequent contextual
disambiguation (see paragraph \ref{res}).
\end{enumerate}

\subsubsection{Incorporating ontologies}
\label{subsubsec3.1.1}
Ontologies can be divided into two types. On the one hand are domain
ontologies, which are formal representations of the kinds of entities
constituting a given domain of inquiry and of the relations between
such entities \citep{smith:2003}. On the other hand are top-level
ontologies, which represent the categories that are shared across a
maximally broad range of domains ~--~  categories such as \textit{object, property,
process} and so forth. Each ontology is built around a taxonomic
hierarchy in which the types of entities are related to each other by the
relation of greater and lesser generality (an analogue of the subset
relation that holds between the instances of such types). 
Domain ontologies have enjoyed considerable success in the formalisation of
the descriptive content of scientific theories above all in many areas of
biology (see especially the Gene Ontology, \citep{ashburner:2000}), where
they served initially as controlled, structured vocabularies for describing
the many new types of entities discovered  in the wake of the Human Genome
Project. On the other hand are top-level ontologies such as Basic Formal
Ontology (BFO, \citep{arp:2015}), which arose to allow domain ontologies at
lower levels to be created in such a way that they share a common set
of domain-independent categories. 
As more and more such domain ontologies came to be developed and applied
to the annotation and management of more and more different types of
biological and biomedical data, the use of such a common top level allowed
the resultant ontologically enhanced data to be more easily combined and
reasoned over.
BFO is now used in this way as shared top-level ontology
in some 300 ontology initiatives.\footnote{BFO is currently under review as
an International Standards Organization
standard under ISO/IEC: 21838-1 (Top-Level Ontologies: Requirements) and
ISO/IEC: 21838-2 (BFO).}

The use of a common top level also allows multiple
ontologies to facilitate standardised exchange between parties
communicating data about entities in different but overlapping domains. Through the
incorporation of formal definitions, they also allow the application of
basic inference mechanisms when interpreting data exploiting taxonomic and
other relations built into the ontology. For logic-based AI applications,
ontologies are needed which reflect the full spectrum of
language constituents and of their logical counterparts. They must enable
the expression not only of traditional taxonomical and mereological
relations but also, for example, of synonymy relations at both
the word and phrase level.

\paragraph{Resolving ambiguity}\label{res}
The terms in such ontologies are defined using formulae of FOL and
the way these formulae represent language can be illustrated using the
second uncertainty example we introduced in section \ref{unc} above.\\

\begin{centering}
\textit{Natural language input: 
John's father did not return. Now John is searching for him.\\
\textit{Formal representation:} $
\textrm{father}(x) \land \textrm{john}(y) \land \textrm{mod}(x,y) \land \lnot
\textrm{Return}_p(x) \land \textrm{Searches}_i(x,y)$}\\
\end{centering}
\bigskip
\noindent Here unary predicates (nouns) and modifiers (mod) indicating
syntactical relationships (in this case: possessive)
are shown in lower case; non-unary predicates (verbs) are shown in upper case,
and the subscript
$_p$ indicates past tense. The subscript $_i$ indicates intensionality of
the transitive verb, which is obtained from the temporal relationship of the two
sentences and the fact that the presence-inducing verb of the
first sentence is negated. The intensionality indicator
might be used in the next step as disambiguation anchor, triggering a search
in the subsequent text for a fulfilling object of the intensional predicate.

\subsection{The core of the philosophy-driven machinery}

In Appendix \ref{section:A}, we give a detailed example of how our approach
in building real-world AI systems combines automated transformation of text into logic with
deterministic and stochastic models. In what follows,
we describe the core functionality we developed to arrive automatically at
a computational text representation, using logic that is semantically
faithful to the input text.

\subsubsection{Transforming text into logic} \label{pipe}
The process of transforming text into logic starts with stochastic
error correction and syntactical tagging using dNN-Part-Of-Speach-taggers
\citep{spacy:2018}. This output is used to perform a sequence of
inferences, starting with:

\begin{equation}
{\textrm{text}} \leadsto \Gamma
\end{equation}

\noindent where `text' is, for example, a sentence from a customer letter;
$\leadsto$ means automated translation; and $\Gamma$ is the set of
logic formulae\footnote{This is a non-compact and non-complete k-order
intensional logic;  `k-order' means that predicates
of the logic can predicate over other predicates arbitrarily often.
`Intensional' means that the range of predication in the logic is not
restricted to existing entities \citep{gamut:1991}.} generated by the
translation. The formulae in $\Gamma$ are generated with a
proprietary AI-algorithm chain that uses world knowledge in the form of a dictionary of
lexemes and their morphemes along with associated rules, relating, for example, to the transitivity and
intensionality of verbs.

\begin{equation}
\Gamma \curvearrowright \Delta
\end{equation}

\noindent where $\Delta$ is a collection of (first-order or propositional
modal) logical formulae automatically generated ($\curvearrowright$) from $\Gamma$. This
action transforms the non-computable formulae of $\Gamma$ into
formulae expressed in logical dialects each of which
enjoys compactness and completeness. The translation
from $\Gamma$ to $\Delta$ requires world knowledge, for example about temporal
succession, which is stored in the computer using ontologies.

\begin{equation}
\Delta \vdash \phi_i \in \Omega, \forall i=1 \dots n 
\end{equation}

\noindent where $\vdash$ means: entailment using mechanical theorem proving, and
$\phi_i$ is one of $n$ human-authored domain-specific formulae $\phi$ entailed by
$\Delta$.

Unlike the automatically generated collections $\Gamma$ and $\Delta$,
$\Omega$ is an ontology comprising human-authored domain formulae $\phi_i$.
$\Omega$ is always related to a specific type of text (for instance, repair
bills) and to a pertinent context (for instance, the regulations under which
the repair occurs). The role of $\Omega$ is to express the target semantics
that can be attributed to input sentences of the given type and context.
$\Delta \cap \Omega \neq  \varnothing$ (i.e. we can infer some $\phi_i$
from $\Delta$) holds only if the input text matches
the type and context of the ontology.

In total, the process looks like this:

$$
\textrm{text} \leadsto \Gamma \curvearrowright \Delta \vdash \phi_i \in
\Omega, \forall i=1 \dots n,
$$

\noindent where the only manual input is the creation of $\Omega$, and
where this manual input itself needs to be performed only once, at system design time.

The Appendix below describes one example of how this approach is embedded
already in a real-world AI production system.

\section{Conclusion}
As becomes clear from the example given in the Appendix below,
our approach to philosophy-driven language AI is to
generate a specific system for each application domain.
There thus remains very little similarity to the hypothetical idea
of \textit{general artificial intelligence.} What we have is rather an exact,
philosophy-driven context- and task-specific \textit{AI technology}.
Systems based on this technology are being successfully used in a range of
different domains. Moreover, the method in question is generalizable to
data of many different sorts, in principle ~--~  as the breadth of the
available ontologies is extended and the sophistication of the algorithms
is enhanced ~--~  covering more and more areas and domains of
repetitive work of the sort amenable to automation. The pace of this
extension to new domains will be
accelerated by enhanced ontology authoring software 
as well as by support for semi-automated ontology generation, for example using
inductive logic programming \citep{ilp:2008}. This will allow for
applications such as automated encoding of medical discharge summaries,
validation of the medical necessity of diagnostic and therapeutic
procedures, and automation of customer correspondence.

We believe that these developments have implications
beyond the merely technical (and, associated therewith, pecuniary). For
they point to a new conception of the role of philosophy in human affairs
which has been evolving since the end of the nineteenth century.

Beginning with the mathematician-philosopher Gottlob Frege, philosophers
have been developing the methods
which enable the expression in exact logical form of knowledge otherwise
expressed in natural language. FOL itself was invented by Frege in 1879,
and since then the FOL framework has been refined and extended to the point
where it is possible to represent natural language in a
formal, computable manner.\footnote{An overview is given in
\citet{boolos:2007}.}

Philosophers have, from the very beginning, attempted to understand
how human language works and how language relates to the world. (Think of
Aristotle's \textit{Organon} and Book VII of his \textit{Metaphysics}.) In the 20th century,
an entire branch of the discipline ~--~  called `analytical philosophy' ~--~  has
grown up around this topic \citep{dummett:1996}. 
The computational discipline of `formal ontology', has in recent years achieved
considerable maturity in part as a result of the influence of philosophical
ideas.

In spite of all this, however, there are many, especially in the twentieth
century, who have proclaimed the death of philosophy, or who have seen
philosophy as having a merely compensatory role in offering some sort of
substitute for those traditions which, in former times, gave human beings
the ability to interpret their lives as meaningful. The ways in which human
lives are meaningful ~--~  are indeed full of meaning ~--~  did indeed play a role in our
argument above. But we would like to draw a more far-reaching conclusion from this argument,
drawing on the ways in which, beginning
already with the Greeks, philosophers have helped to lay the groundwork for
a series of social upheavals in the course of human history.
These include, for example, the birth of
democracy or of market institutions, of new artefacts such as Cartesian
coordinates, and even of entire scientific disciplines. For we believe
that one place where we can look for a role for
philosophy in the future will lie in the way it can be used to strengthen
and enable applied sciences in the
digital era ~--~ for example, in the creation of useful and realistic
artificial intelligence applications involving automatic translation of
natural language texts into computer-processable logical formulae.

\appendix

\section{Appendix: A real-world example}
\label{section:A}

To represent in logical form the full meaning of complex natural
language expression E as used in a given domain and for a given purpose, we will
need a set of domain-specific ontologies together with algorithms which,
given E, can generate a logical
formula using ontology terms which are counterparts
of the constituent simple expressions in E and which expresses the
relations between these terms. These algorithms should
then allow the representation in machine-readable form not merely of single expressions
but of entire texts, even of entire corpora of texts, in which
domain-specific knowledge is communicated in natural language form.

To see how philosophy is already enabling applied science-based production along these
lines, let us look at a real-world example of an AI automaton used to
automatically generate expert technical appraisals for insurance
claims.\footnote{\label{fn2}This software is in production at 
carexpert GmbH, Walluf, Germany, a claims validation service provider
processing some 70 thousand automobile glass repair bills a year with a
total reimbursement sum of over \EURtm 50 million.  The bill validation
process, performed by a car mechanics expert in 7-10 minutes, is completed
by the system in 250 milliseconds.}
Today, such claims are validated by mid-level clerical staff, whose job is to
compare the content of each claim ~--~  for example the line items in a car
repair or cardiologist bill ~--~  with the standards legally and technically
valid for the context at issue (also referred to as `benchmarks').
When deviations from a benchmark are detected
by humans, corresponding amounts are subtracted from the indemnity amount with a
written justification for the reduction. Digitalization has advanced
sufficiently far in the insurance world that claims data can be made
available in structured digital form (lines in the bill are stored as
separate attributes in a table in a relational database). However, 
the relevant standards specifying benchmarks and how they are to be
treated in claims processing have until recently
been represented only as free-text strings. Now, however, by using
technology along the lines described above, it is possible to automate both
the digital representation of these standards and the results of
the corresponding comparisons between standards and claims data.

To this end, we developed an application that combines stochastic models
with a multi-facetted version of logic-based AI to achieve the following
steps:

\begin{enumerate}
\item	Compute a mathematical representation (vector) of the contents of the
bill using logic for both textual and quantitative data.
The text is vectorised using a procedure shown 
in equations (1) -- (3) of section \ref{pipe} above, while
the quantitative content is simply inserted into the vector.
\item	Recognise the exact type of bill and understand the context in
which it was generated. This is done using the logical representation of
the text, which is taken as input for deterministic or stochastic
classification of the bill type (for example, \textit{car glass damage}) and
subtype (for example, \textit{rear window}).
\item	Identify the appropriate repair instructions (`benchmark') for the bill by
querying the corresponding claims knowledge base for the benchmark
most closely matching the bill in question. Standard sets of benchmarks are provided by the original
equipment manufacturers or they are created from historic
bills using unsupervised pattern identification in combination with
human curation. The benchmark texts are transformed into mathematical logic.
\item	Compare the bill to its benchmark by identifying matching 
lines using combinatorial optimisation. The matches are
established by
computing the logical equivalence of the matching line items using
entailment in both directions: given a bill line (or line group) $p$ and
its candidate match (or group) $q$,
compute $p \vdash q$ and $q \vdash p$ to establish the match.
\item	Subtract the value of the items on the bill that do not match the
benchmark from the reimbursement sum
\item	Output the justification for the subtractions using textual
formulations from the appropriate standard documents.
\end{enumerate}

To achieve comparable  results an end-to-end dNN-based algorithm would
require billions of bills with standardised appraisal results. Yet the
entire German car market yields only some 2-3 million car glass damage repair
bills in any given year and the appraisals are not standardised.

The technology is used for the automation of typical mid-level office tasks. It
detects non-processable input, for example language resulting in 
a non-resolvable set of logical formulae, and passes on the cases it cannot process
for human inspection. This is a
core feature of our technology which may not match the expectations of an AI
purist. However, applications of the sort described
have the potential to automate millions of office jobs in the
German-speaking countries alone.

Human beings, when properly trained, are able to perform the classification
described under step 2 spontaneously. They can do this both for entire
artefacts such as bills and for the single
lines which are their constituents. Humans
live in a world which is meaningful in precisely this
respect.\footnote{Compare footnote \ref{fn} above.} The ability
to classify types of entities in given contexts can be replicated in
machines only if they store a machine-adequate representation of the background
knowledge that humans use in guiding their actions. This is realised in the
described system by means of ontologies covering both the entities to which reference is made in given
textual inputs and the contexts and information artefacts associated therewith. The ontologies
also incorporate formal definitions of the relevant characteristics of these objects, of the
terms used in the relevant insurance rules, and so forth. The ontologies
are built by hand, but involve a minimal amount of effort for those with expertise in
the relevant domain (here: the contents of repair bills and insurance rules).
These definitions are entailed by the bill and benchmark texts, and the
latter are automatically processed into logical representations in the
ontology framework without human interference.

\subsection*{Resulting system properties}

This philosophy-driven AI application uses both stochastic models and
parsers, as well as mechanical theorem provers. It meets the requirements
listed in Table 1, including:

\begin{itemize}
\item \textit{Exactness}  ~--~  it has an error rate of below 0.3\% (relative
to the gold standard obtained by a consortium of human experts), which is
below the best human
error rate of 0.5\%. Such low levels of error are achieved only because,
unlike a stand-alone stochastic model, the system will detect if
it cannot perform any of the essential inference steps
and route the case to a human being.
\item \textit{Information security}  ~--~  the system is secure because any
misreactions to perturbing input by its stochastic models are
detected by the logical model working in the immediately subsequent step.
\item \textit{Robustness}  ~--~  it is robust since it will detect when
it cannot interpret a given context properly, and issue a corresponding
alert.
\item \textit{Data parsimony} ~--~  it requires very little data for
training, since unlike the sorts of suboptimally separating agnostic spaces resulting from stochastic embeddings,
it induces what we can call a semantic space that separates data
points very effectively.
\item \textit{Semantic fidelity}  ~--~  the system not only allows,
but it is in fact based on inference and so it can easily use prior and world
knowledge in both stochastic (Bayesian net) and deterministic (logical) form.
\end{itemize}

\subsection*{Acknowledgements}
We would like to thank Prodromos Kolyvakis, Kevin Keane, James Llinas and
Kirsten Gather for helpful comments.

\printbibliography


\end{document}